\documentclass[11pt,a4paper]{article}
\usepackage[hyperref]{acl2021}
\usepackage{times}
\usepackage{latexsym}

\usepackage{amsfonts}
\usepackage{bbold}
\usepackage[T1]{fontenc}

\usepackage{microtype}

\aclfinalcopy 

\title{\sv: Interactive Visual Analysis of Models, Data, and Evaluation for Text Summarization}

\author{
  \textbf{Jesse Vig}\textsuperscript{$\ddagger$}
  \quad \textbf{Wojciech Kry\'sci\'nski}\textsuperscript{$\ddagger$}
  \quad \textbf{Karan Goel}\textsuperscript{$\dagger$}
  \quad \textbf{Nazneen Fatema Rajani}\textsuperscript{$\ddagger$} \\
  \textsuperscript{$\ddagger$} Salesforce Research
  \quad 
  \textsuperscript{$\dagger$} Stanford University \\
  \texttt{\{jvig, kryscinski, nazneen.rajani\}@salesforce.com} \\
  \texttt{kgoel@cs.stanford.edu} \\ 
}

\date{}

\newcommand{\sv}{{\sc{SummVis}}}

\usepackage{graphicx}
\usepackage{subfigure}
\usepackage{enumitem}
\usepackage{float}
\usepackage{color,soul}
\usepackage[normalem]{ulem} 
\newcommand{\TriangleFigure}{
\begin{figure}[t]
    \centering
    \includegraphics[width=\linewidth]{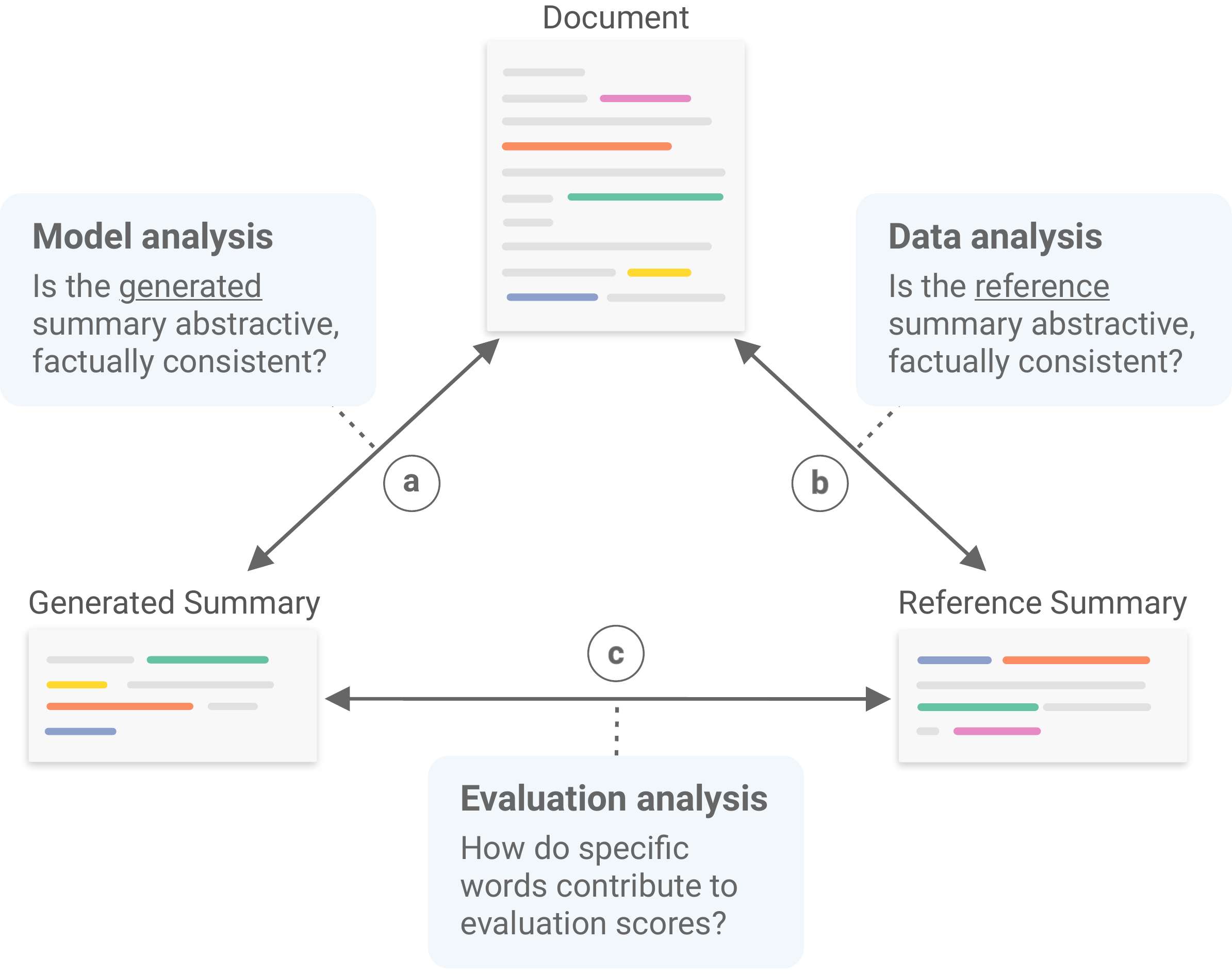}
    \caption{\sv{} supports fine-grained comparison between (a) source document and generated summary, (b) source document and reference summary,  and (c) generated summary and reference summary, enabling analysis of models, data, and evaluation metrics.}
    \label{fig:summvis-analysis-modes}
\end{figure}
}

\newcommand{\InterfaceFigure}{
\begin{figure*}[t!]
    \centering
    \includegraphics[width=1\linewidth]{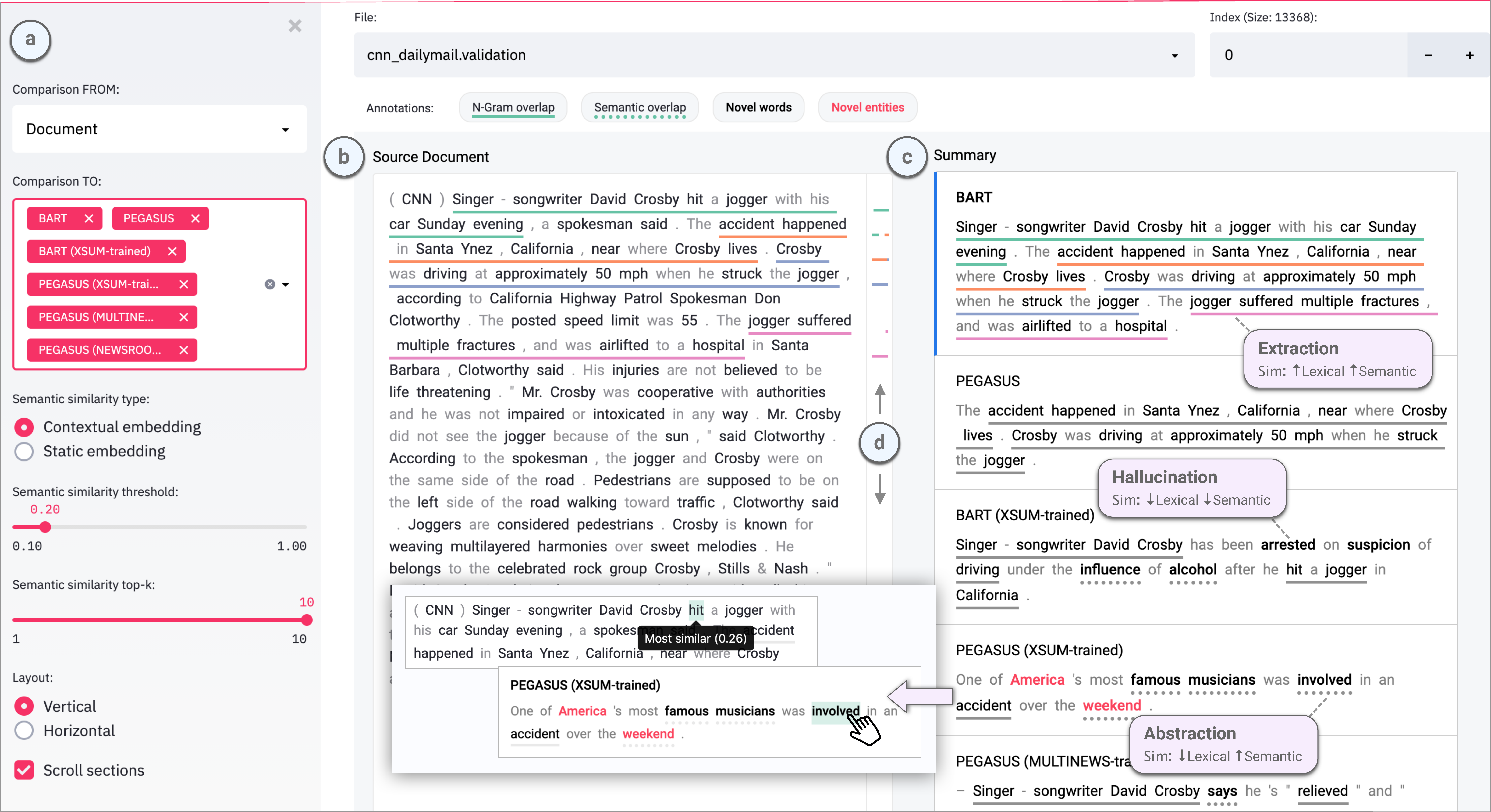}

    \caption{\textbf{Warning: Contains factually incorrect information.} \sv{}~interface showing the first example from the CNN/DailyMail validation split. Interface components: (a) configuration panel, (b) source document (or reference summary, depending on configuration), (c) generated summaries (and/or reference summary, depending on configuration), (d) scroll bar with global view of annotations. \setulcolor{green}\ul{Colored} \setulcolor{orange}\ul{underlines} align n-grams between source document and the selected summary (BART); colors are determined by the position of the containing sentence in the summary. Novel words in the summary that do not appear in the source document are \textbf{bolded}, while novel entities are bolded in {\color{red}\textbf{red}}. Stopwords are {\color{gray}grayed out} and are not used in the matching algorithms. \dotuline{Dotted underlines} indicate tokens that are semantically similar to a token in the source document (above the threshold specified in the configuration panel). The user may hover over a token to see the most semantically similar tokens in the source document (see inset image), or click on the token to auto-scroll the source document to the most similar token.} 
    \label{fig:interface-main}
    \vspace{10pt}
\end{figure*}
}

\newcommand{\CaseStudyInterface}{
\begin{figure*}[t!]
    \centering
    \includegraphics[width=1\linewidth]{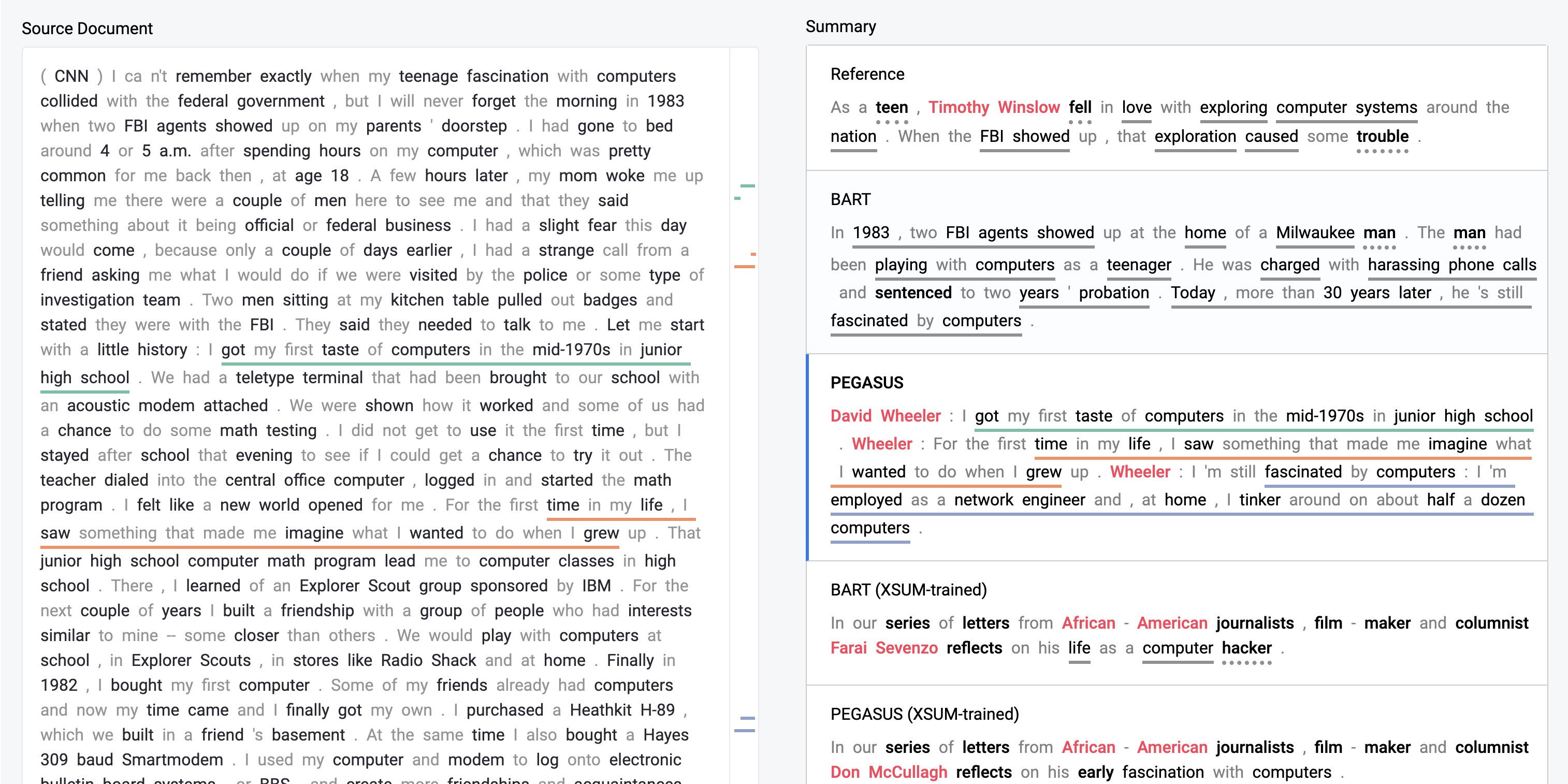}
  \caption{\sv~snapshot of an example from CNN/DailyMail, showing the source document (left) and the reference summary along with generated summaries from four different models (right). The first two models are trained on CNN/DailyMail, while the last two are trained on XSum. {\color{red}\textbf{Red}} text  highlights entities in the summaries that are not present in the source document.}
    \label{fig:case-study-interface}
\end{figure*}
}

\begin{document}
\maketitle
\begin{abstract}
Novel neural architectures, training strategies, and the availability of large-scale corpora haven been the driving force behind recent progress in abstractive text summarization.
However, due to the black-box nature of neural models, uninformative evaluation metrics, and scarce tooling for model and data analysis, the true performance and failure modes of summarization models remain largely unknown.
To address this limitation, we introduce \sv{}, an open-source tool for visualizing abstractive summaries that enables fine-grained analysis of the models, data, and evaluation metrics associated with text summarization. 
Through its lexical and semantic visualizations, the tools offers an easy entry point for in-depth model prediction exploration across important dimensions such as factual consistency or abstractiveness.
The tool together with several pre-computed model outputs is available at~\url{https://github.com/robustness-gym/summvis}.
\end{abstract}
\section{Introduction}\label{sec:introduction}

The field of Natural Language Processing has seen substantial progress in recent years driven by the availability of large-scale corpora~\cite{Brown:20, Raffel:20}, developments in neural architectures~\citep{Vaswani:17, Zaheer:20} and training strategies~\cite{Devlin:19, Zhang:20}.
Despite the promising results on benchmarks and recent findings in model analysis, the true performance, generalizability, and failure modes of modern neural models are not yet fully understood, due to the black-box nature of neural models and the unmanageable scale of recent datasets for manual analysis.
Software tooling for NLP research provides a plethora of mature and easy-to-use libraries for model development, such as PyTorch~\citep{Paszke:19} or Transformers~\citep{Wolf:20}, but offers disproportionately fewer tools for visual analysis and debugging, which further hinders the understanding of model performance.

\TriangleFigure
\InterfaceFigure
Within NLP, Automatic Text Summarization is a task that aims to convert long documents into short textual snippets that contain the most important information from the source document.
To successfully summarize documents, models must first build an understanding of the source text that will allow them to evaluate the saliency of presented facts and then select only the most important details for the output summary. 
In case of abstractive approaches, the neural networks are also expected to paraphrase the selected content to generate novel sentences that fuse together the facts extracted from different sections of the document into coherent and factually consistent text.

Progress in the field is measured primarily using automatic metrics, such as  ROUGE~\citep{Lin:04} or BERTScore~\citep{Zhang:20b}, which quantify the lexical and semantic overlap between reference and generated summaries.
While automatic metrics are convenient for model evaluation, they have been shown to be mismatched with human judgements~\citep{Fabbri:20} and only offer high-level insights while failing to pinpoint particular shortcomings of models.
In-depth debugging across the different modes of analysis (Fig.~\ref{fig:summvis-analysis-modes}) must be conducted through expensive and time-consuming human-based studies, where the substantial length of texts makes such efforts more labor-intensive.

Recent work in summarization analysis has looked at the problems of the field in isolation, focusing on: models~\citep{DBLP:conf/emnlp/KedzieMD18, Kryscinski:19, Kryscinski:20}, data~\citep{zhong2019closer,jung2019earlier}, and evaluation~\citep{Fabbri:20,steen2021evaluate}.
However, these modes of analysis are strongly interconnected and isolating them could skew the broader view of the current state of the task and delay progress.

To address the mentioned challenges, we introduce \sv{}, an open-source interactive visualization tool for analyzing text summarization.
\sv{} was designed to offer fine-grained insights into the models, data, and evaluation metrics, both in isolation and jointly, thus compensating for the shortcomings of automatic evaluation metrics and shortage of dedicated debugging tooling.
\sv{} scaffolds human analysis by offering clear visual indicators of the semantic and lexical relationships between texts and intelligent navigation within text.
The tool comes pre-loaded with a set of state-of-the-art model predictions for a quick starting point for model analysis and comparison and offers out-of-the-box integration with the HuggingFace Dataset API for custom use-cases.
Through a case study of state-of-the-art summarization models we show how \sv{} can be used to quickly conduct non-trivial analysis, debugging, and comparison of model performance across important dimensions such as factual consistency or abstractiveness. 
%

\section{\sv}\label{sec:summviz}
In this section, we present \sv{}, an interactive visualization tool that provides rich text comparison in summarization systems, enabling fine-grained analysis of models, data, and evaluation metrics. It comes pre-loaded with model outputs for state-of-the-art models over common benchmark datasets, as well as scripts for loading data for any dataset provided by the Datasets API~\citep{Wolf:20a} and any HuggingFace-compatible model.

\subsection{Analysis Modes}
\label{sec:analysis_modes}
\sv{}~supports three modes of analysis, depending on the type of text being compared: 

\begin{enumerate}[leftmargin=*]
        \item {\bf Model Analysis} (Fig.~\ref{fig:summvis-analysis-modes}a). By comparing the source document with \textit{generated} summaries, \sv{} provides insights into a model's ability to abstract and faithfully retain information present in the document.
        \item {\bf Data Analysis} (Fig.~\ref{fig:summvis-analysis-modes}b).  By comparing the source document with the \emph{reference} summary, \sv{} helps determine the degree to which the reference summary itself is abstractive and factually consistent with the source document.
        \item {\bf Evaluation Analysis} (Fig.~\ref{fig:summvis-analysis-modes}c). By comparing the \emph{reference} summary with the \emph{generated} summary, \sv{} surfaces the word- and phrase-level relationships that form the basis of automated evaluation metrics such as ROUGE and BERTScore.
\end{enumerate}

These analyses are interdependent with one another; for example, the behavior of a \textit{model} depends on the \textit{data} on which it was trained. By providing a unified interface for all modes of analyses, the user may also draw conclusions about the relationships between model, data, and evaluation, as we'll demonstrate in Section~\ref{sec:case-study}.

\subsection{Text Comparison}
\label{sec:comparison}
Understanding abstractive summaries requires comparing not only surface similarities but also building a semantic understanding of the source document and summaries. Therefore~\sv{} incorporates similarity measures based on both lexical and semantic overlap, as described below.

\noindent{\bf Lexical Overlap.} The ability to quickly compare the lexical form of source document and summary is an important first step in analyzing a generated summary. For example, it is well known that many abstractive reference summaries are in fact largely extractive, copying long spans of text from the source document~\citep{grusky2018newsroom}. Other summaries might contain significant hallucinations, including words that are not found in the source document~\citep{Kryscinski:20,maynez2020faithfulness}. In order to identify these phenomena, \sv{} provides a lexical alignment based on shared n-grams between the two texts, which is also the basis for many automated metrics such as ROUGE.

\noindent{\bf Semantic Overlap.} Lexical overlap is incomplete as a measure of similarity between texts since it only considers the surface form of words. For example, a summary that is highly abstractive may share few common words with the source article, despite having a similar meaning. To address such limitations, the tool also identifies semantically-related tokens by computing the cosine similarity between word embeddings, with the option of using static word embeddings provided by spaCy~\citep{spacy:20}, or contextual embeddings from a pretrained RoBERTa~\citep{liu2019roberta} model. In the later case, we apply the same default embeddings\footnote{RoBERTa-large layer 17} used in BERTScore, a common evaluation metric for abstractive summarization systems that correlates strongly with human evaluations~\citep{Zhang:20b}.  As we'll discuss in Section~\ref{sec:case-study}, the visualized semantic similarities can also help to interpret BERTScore values. We note that the BERTScore library\footnote{https://github.com/Tiiiger/bert\_score} used in the tool also supports other models of semantic similarity, for example, models trained on scientific or non-English text.

\begin{figure}[t!]
    \centering
    \includegraphics[width=.85\linewidth]{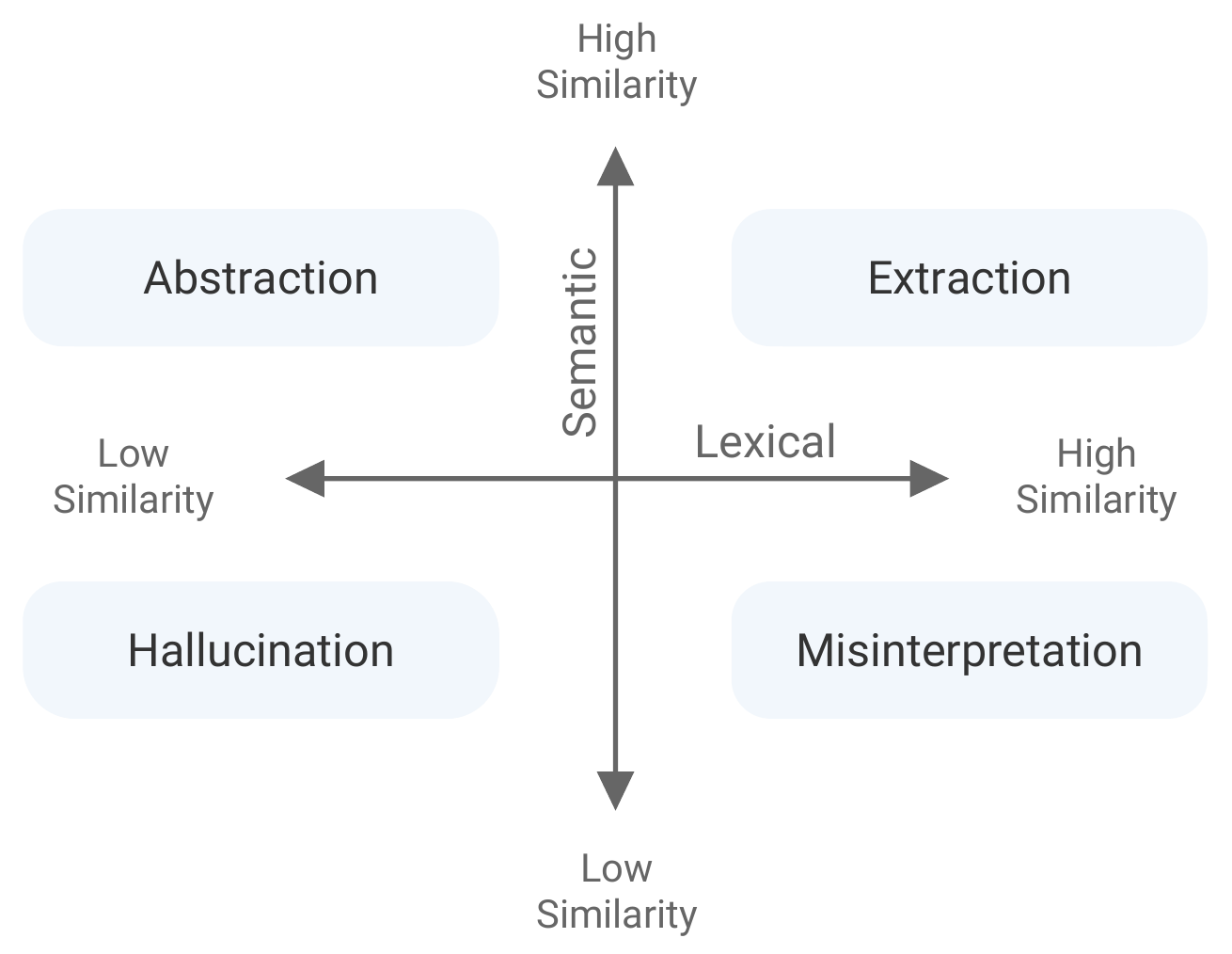}
    \caption{Taxonomy of textual relationships across lexical and semantic dimensions.}
    \label{fig:summvis-taxonomy}
\end{figure}

\noindent{\bf Taxonomy.} Considering both lexical and semantic measures of similarity provides a natural way to chart out summarization datasets for further analysis. By comparing a source document to any summary along these two dimensions, four quadrants of behavior can be mapped out (Fig.~\ref{fig:summvis-taxonomy}):
\vspace{-.4em}
\begin{enumerate}[leftmargin=*]
  \setlength\itemsep{-.2em}
    \item {\bf Extraction}: high lexical and high semantic similarity. The summary quotes text from the document verbatim.
    \item {\bf Abstraction}: low lexical and high semantic similarity. The summary consolidates and paraphrases information from the document.
    \item {\bf Hallucination}: low lexical and low semantic similarity. The summary is factually inconsistent, and includes information that is absent in the document.
    \item {\bf Misinterpretation}: high lexical and low semantic similarity. The summary misinterprets and uses information from the document, such as misunderstanding homonyms.
\end{enumerate}

Examples of such cases will be discussed in the following sections. 

\subsection{Interface}
The main components of the \sv{} interface are described in detail in Figure~\ref{fig:interface-main}. The interface supports analysis of the model, data, and evaluation (Sec.~\ref{sec:analysis_modes}) based on which types of text are selected by the user for comparison (Figs.~\ref{fig:interface-main}b, \ref{fig:interface-main}c). 
The annotations provided by the tool highlight both lexical and semantic relations between the text (Sec.~\ref{sec:comparison}) and are designed to be lightweight, allowing users to quickly grasp the relationship between texts while still being able to clearly read the text. 

The joint lexical and semantic annotations enable the user to understand the summaries according to the taxonomy in Figure~\ref{fig:summvis-taxonomy}. Examples of extraction, abstraction, and hallucination are highlighted in Figure~\ref{fig:interface-main}. Since measures of semantic similarity may be unreliable, the tool also enables users to hover over tokens for additional details on the semantically matched tokens in the source document, which are highlighted based on their semantic similarity scores (Fig.~\ref{fig:interface-main}, inset image). Additionally the score of the closest match is displayed, following the BERTScore algorithm, which computes the maximum semantic similarity score for each token before averaging the results over the full text. These features enable users to manually assess whether the tokens are in fact semantically similar. 

The tool supports two additional features to accommodate long source documents: a global view and auto-scrolling functionality. The global view, embedded in the scroll bar region of the source document (Fig.~\ref{fig:interface-main}d), displays a compressed view of the full document's annotations that is visible even when the document exceeds the viewable region. The user may also directly navigate to matched portions of the source documents not currently visible by clicking on related annotations in the summary. 

\subsection{System Architecture}

The interface is implemented as a Streamlit\footnote{https://github.com/streamlit/streamlit} application with a highly customized HTML/JavaScript component that handles most interactions in the tool. The custom component enables a much richer interaction than a vanilla Streamlit app, while the Streamlit infrastructure allows for adapting or extending some components in the tool without necessarily writing additional HTML or JavaScript.

We provide pre-processing scripts to generate and cache all data required by \sv{} to ensure fast response times in the interface. These scripts are implemented using Robustness Gym~\citep{goel2021robustness} and integrate with the HuggingFace Datasets API~\citep{Wolf:20a} so that any summarization dataset available in the dataset repository or provided by the user as a \texttt{jsonl} file may be viewed in the tool. We additionally include scripts for caching outputs for any HuggingFace summarization model, and share pre-computed outputs of state-of-the-art summarization models: PEGASUS~\citep{Zhang:20} and BART~\citep{Lewis:19}.
To increase the variaty of outputs, we chose model checkpoints fine-tuned on multiple popular summarization datasets: CNN/DailyMail~\citep{Hermann:15}, XSum~\citep{Narayan:18}, Newsroom~\citep{grusky2018newsroom}, and MultiNews~\citep{fabbri-etal-2019-multi}, and decoded on the validation splits of two benchmark datasets: CNN/DailyMail and XSum.
\section{Case Study: Debugging Hallucination}\label{sec:case-study}

\CaseStudyInterface

\begin{figure*}
  \centering
    \includegraphics[width=\linewidth]{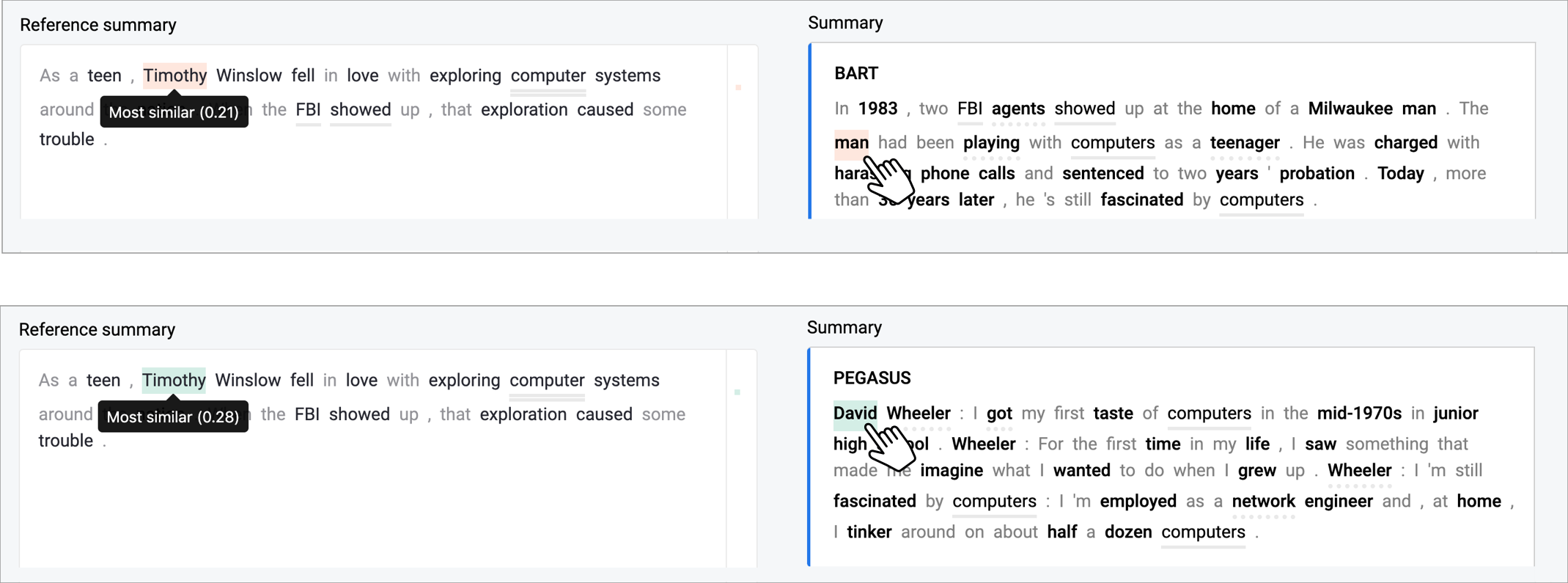}
  \caption{Snapshot from \sv~showing the reference summary on the left and two of the generated summaries on the right. In the first example, the user has hovered over ``\textit{man}'' in the generated summary, which causes the tool to highlight the most semantically similar word in the reference summary, ``\textit{Timothy}'', with a similarity score of 0.21. A second occurrence of ``\textit{man}'' has an even lower semantic similarity score of just 0.02 (not shown). In the second example, the user hovers over ``\textit{David}'', revealing that this word is also most semantically similar to ``\textit{Timothy}'', but with a higher similarity score (0.28).}
  \label{fig:case-study-ref}
\end{figure*}

As discussed earlier, \sv~ supports joint analysis of the model, data, and evaluation metrics. We now demonstrate how we can draw from all three modes of analysis to study the problem of hallucination in summarization systems. Through the unified view of \sv, we analyze the example shown in Figure~\ref{fig:case-study-interface} and demonstrate the existence of hallucination, suggest a possible cause, and show how a common evaluation metric prefers hallucinated entities over faithful descriptors in this case.

\paragraph{Model Analysis.} \sv{} supports analysis of the model by visualizing the relationship between each generated summary and the source document. For the example in question (Fig.~\ref{fig:case-study-interface}), this visualization reveals that three of the four models generate names of people that are absent from the source document. The XSum-trained models generate the names in the context of the phrase \textit{``In our series of letters from African-American journalists, film-maker and columnist <person\_name> reflects on ...''}. suggesting that the hallucinations for these two models may be related to artifacts in the shared XSum training set that both models have memorized. 
On the other hand, the summary generated by the version of PEGASUS that was trained on CNN/DailyMail is largely extractive, copying several sentences, but then also inserting the name ``\textit{David Wheeler}'', which is absent from the source document.  We now show how artifacts in the \textit{reference} summaries may explain this hallucination.

\paragraph{Data Analysis.} We now turn to the visualization comparing source document and \textit{reference} summary (Fig.~\ref{fig:case-study-interface}, top right). We see that the reference summary also contains an entity that is missing from the source document (``\textit{Timothy Winslow}''). This may be due to the name appearing in metadata such as author name that was available to the person writing the summary, but was not included in the dataset.  If this pattern occurs in similar types of examples in the training set (e.g., first-person written articles), then it may effectively teach the model to hallucinate, providing a possible explanation for the model behavior described earlier.

\paragraph{Evaluation Analysis.}

One remaining question is how state-of-the-art models can hallucinate but still perform well on benchmark datasets according to standard evaluation metrics. Of course, one reason is that the models only hallucinate on some fraction of examples in the dataset. However, there is also the question of how the evaluation metrics score hallucinated content. While lexical overlap metrics such as ROUGE are well-defined, semantic similarity metrics like BERTScore are less well understood as they depend on embeddings from black-box neural network models.

\sv{} supports fine-grained analysis of evaluation metrics through its comparison of \textit{generated} and \textit{reference} summaries. In particular, the token-level semantic similarity scores visualized in the tool use the same similarity measure as BERTScore (Sec.~\ref{sec:comparison}). By inspecting these token-level relationships, we can better understand how hallucinated tokens contribute to the overall BERTScore, which is computed by aggregating token-level scores. 

Figure~\ref{fig:case-study-ref} shows the comparison between the reference summary and two of the generated summaries, revealing that the factually correct ``\textit{man}'' has a lower maximum semantic similarity score compared to the hallucinated ``\textit{David}''. The same is true for the corresponding hallucinated last name ``\textit{Wheeler}'' (similarity: 0.28), and this disparity with ``man'' is even more pronounced for the hallucinated name ``\textit{Don McCullagh}'' (Similarity:  0.34, 0.31) generated by the last model shown in Figure~\ref{fig:case-study-interface}. Thus BERTScore does not discriminate factual consistency of proper names in this example,  consistent with anecdotal evidence for other types of entities~\citep{Zhang:20b}. Note that the hallucinated name ``\textit{Farai Sevenzo}'' (Fig.~\ref{fig:case-study-interface}, 4th row) has maximum similarity scores that are negative (-0.43, -0.12). This disparity may relate to name biases in word embeddings~\cite{Caliskan183}. 

\section{Related Work}\label{sec:related-work}
Text Summarization requires models to be adept at both natural language understanding (NLU) and natural language generation (NLG).
A gap in either of these areas has consequences on the progress of summarization as a whole.
An example of this is the lack of meaningful metrics in NLG for high entropy tasks~\citep{steen2021evaluate}.
Several recent works have realized the need for evolving benchmarks and evaluations~\citep{goel2021robustness,gehrmann2021gem,khashabi2021genie}.

Existing tools support some forms of text comparison for summarization models. The Newsroom dataset visualization tool~\citep{grusky2018newsroom} highlights n-grams in the summary that  overlap with the source article. The LIT tool~\citep{tenney2020language} highlights words or characters that differ between reference and generated texts. However neither tool aligns~\citep{9220137} the matched text. The CSI framework~\citep{gehrmann2019visual} and the Seq2SeqVis~\citep{seq2seqvisv1} tool align the source document and summary, but use model-specific attention mechanisms. \mbox{\sv{}~}on the other hand supports a model-agnostic comparison between source document, reference summary, and generated summary, and aligns text along lexical and semantic dimensions.

\section{Conclusion}\label{sec:conclusions}
In this work we introduced~\sv{}, an interactive visualization tool for analyzing text summarization models, datasets, and evaluation metrics.
Through a case study we showed that our tool can be used to efficiently identify the shortcomings and failure modes of state-of-the-art summarization models and datasets.
Together with the tool we released a set of pre-computed model outputs to enable easy, out-of-the-box use.
We hope this work will positively contribute to the ongoing efforts in building tools for model evaluation and analysis and enable a deeper understanding of the performance of summarization models and the intricacies of datasets and metrics.
\section{Ethics Statement}
To the best of our knowledge, there is no work on ethical bias in automated text summarization. The news summarization datasets currently used by the NLP community are mainly crawled from Western news outlets and therefore are not representative of a majority of geographies. There are also biases in news reporting that can distill into parameters of models trained on such biased datasets and may even be further amplified in the generated model outputs.  All datasets are in English, and all models are trained on English datasets.

\sv~uses spaCy for entity detection and because we did not stress test the detector, there might be biases in the system that have percolated into our tool. Similarly, the text similarity metrics used in our tool including the BERTScore and the word-embeddings carry biases of the data they were trained on. For example, they have been known to have bias associating professions with a particular gender. We request our users to be aware of these ethical issues that might affect their analyses.

\section*{Acknowledgments}
We would like to thank Michael Correll for his insightful feedback.

\bibliographystyle{acl_natbib}
\bibliography{anthology,acl2021}

\end{document}